\documentclass{article}
\usepackage{arxiv}
\usepackage{hyperref}
\usepackage{url}
\usepackage{booktabs}
\usepackage{amsfonts}
\usepackage{nicefrac}
\usepackage{microtype}
\usepackage{graphicx}
\usepackage{natbib}
\usepackage{doi}
\usepackage{tabularx}
\usepackage{amsmath} 
\usepackage{multirow}

\title{An End-to-End Hybrid Framework for Rumour Detection in Low-Resources Algerian Dialect}

\author{ \href{https://orcid.org/0000-0002-3794-844X}{\hspace{1mm}Dihia LANASRI, Fatima BENBAREK}\\
	ATM Mobilis, USTHB \\
	Algiers, Algeria\\	
	\texttt{dihia.lanasri@gmail.com , benbarekfatima1@gmail.com} \\
    \\    
}

\renewcommand{\shorttitle}{Rumor Detection in Algerian Dialect}

\hypersetup{
pdftitle={Rumor Detection in Algerian Dialect},
pdfsubject={NLP, DIALECT},
pdfauthor={D.LANASRI and al},
pdfkeywords={Natural Language Processing, Rumor Detection, Algerian Dialect, Transformers},
}

\begin{document}
\maketitle

\begin{abstract}
The rapid growth of social media has intensified the spread of rumours. This issue is more challenging in the Algerian context due to the informal and code-switched nature of dialectal content, the scarcity of annotated resources, and the limited effectiveness of standard Arabic NLP tools on dialect text.

This paper presents an end-to-end rumour detection hybrid framework for Algerian dialect social media content. We build a domain-specific annotated dataset by combining real social media posts, synthetic data, and the FASSILA corpus, with automatic labeling based on a similarity-based annotation process. A transliteration pipeline is also introduced to generate parallel datasets in Arabic script and Arabizi.

We evaluate multiple approaches, including classical machine learning, deep learning, transformers, and hybrid models. Experimental results show that a hybrid approach combining transformer embeddings with a classical classifier achieves the best performance, reaching an F1-score of 0.84. We also find that domain-specific pre-training is more important than model size, with social media–trained models outperforming larger models trained on formal Arabic corpora.

These results demonstrate the feasibility of rumour detection in low-resource Algerian dialect settings.

\end{abstract}


\section{Introduction}
Social media platforms have become the dominant infrastructure for information dissemination in the digital age. Every second, massive volumes of user-generated content are produced and shared across platforms such as Facebook, YouTube, and Instagram, etc., enabling information to propagate with unprecedented speed and reach. While this connectivity has transformed communication and access to information, it has also amplified the spread of unverified claims, misinformation, and rumours, which can rapidly influence public perception before any factual correction is issued.

In this context, rumour propagation has emerged as a critical challenge across multiple domains, including public services, finance, and telecommunications. In particular, the telecommunications sector is highly sensitive to such phenomena, as false claims regarding network outages, service degradation, pricing policies, or subscription offers can quickly escalate, leading to customer dissatisfaction, loss of trust, and long-term damage to brand reputation. The viral nature of social media further intensifies this effect, making early and automatic detection of rumours an essential requirement for modern digital monitoring systems in companies.

However, detecting rumours in real-world settings is far from trivial, especially in low-resource linguistic environments. The Algerian landscape presents a particularly complex case due to its unique linguistic and socio-cultural characteristics. User-generated content is predominantly written in the Algerian dialect (Darja), an informal spoken variety that lacks standardized grammar and orthography. This dialect is inherently heterogeneous, continuously shaped by regional variation and frequent code-switching between Arabic, French, and Tamazight. In addition, a large portion of online communication is written in Arabizi, a non-standard Latin-script representation of Arabic sounds, often mixed with digits and informal abbreviations.

These linguistic properties introduce significant challenges for Natural Language Processing (NLP) systems. Most existing tools are designed for Modern Standard Arabic (MSA), which is structurally and lexically different from dialectal usage. As a result, their performance degrades significantly when applied to Algerian dialect content. The absence of large-scale annotated datasets, combined with the lack of standardized preprocessing pipelines for Arabizi and dialectal text, further limits progress in this area.

Rumour detection has been widely explored in the NLP literature, particularly for high-resource languages such as English, where early work relied on classical machine learning methods using handcrafted features \cite{castillo2011information, qazvinian2011rumor, zubiaga2017exploiting}, followed by deep learning approaches leveraging sequential models \cite{ma2016detecting}, and more recently transformer-based architectures that capture contextual dependencies \cite{devlin2019bert, zhou2020fake}. In Arabic NLP, research has gradually expanded to include Modern Standard Arabic and, to a lesser extent, certain regional dialects \cite{abdalla2020arabic_nlp_survey, boukil2021arabic_dialect_nlp}. However, most existing studies remain focused on formal or semi-formal text sources, and relatively few address highly informal, code-switched dialectal data \cite{alshalan2020arabic_fake_news}. In particular, the Algerian dialect remains less explored, especially in domain-specific applications such as telecommunications monitoring. Only, few researchers have been interested in the question of rumor detection in Algerian dialect \cite{zakaria2023algerian, abdedaiem2024fassila}.

This gap highlights the need for dedicated datasets, robust preprocessing strategies, and adaptive modeling approaches capable of handling multilingual, noisy, and non-standard social media text. Motivated by this challenge, this work focuses on developing an automated rumour detection framework tailored to Algerian dialect content. The goal is to explore how different modeling paradigms, including classical machine learning, deep learning, transformer-based models, and hybrid architectures, perform under such low-resource and linguistically complex conditions.

The remainder of this paper is organized as follows: Section 2 reviews related work in rumour detection in Arabic and Algerian Dialect contexts, Section 3 presents the detailed approach that have been developed, Section 4 details the conducted experiments and obtained results, Section 5 presents a detailed discussion and analysis and Section 6 concluded the paper.

\section{Related Work}
Research on rumour and fake news detection in Arabic has evolved significantly in recent years, spanning classical machine learning, deep learning, and transformer-based approaches. In this section, we review the most relevant contributions, organized by linguistic coverage: Modern Standard Arabic (MSA), other Arabic dialects, and Algerian dialect specifically. Table~\ref{tab:related_work} summarizes the reviewed studies.

\subsection{Modern Standard Arabic (MSA) and Mixed Arabic Settings}
~\cite{zubiaga2018detection} provide foundational work on contextual rumour detection using sequential and neural architectures, which inspired later adaptations to Arabic dialectal settings.

Early work in Arabic fake news and rumour detection primarily focused on Modern Standard Arabic and mixed Arabic datasets. ~\cite{alkhair2019arabic} introduced one of the earliest datasets combining MSA with limited dialectal content, including Algerian and Egyptian texts collected from news and satirical sources. They evaluated classical classifiers such as SVM, Naive Bayes, and Logistic Regression using TF-IDF features, with SVM achieving competitive performance (94\%--96\% accuracy). However, the dataset remains largely MSA-oriented and does not reflect the complexity of social media dialectal usage.

~\cite{hocini2024detecting} proposed a multilingual fake news dataset (BOUTEF) covering MSA, Algerian, Tunisian dialects, and code-switched text. Using keyword-based TF-IDF representations and classical models, they achieved up to 82.3\% accuracy with MLP. Despite its multilingual nature, the reliance on predefined keyword lists limits contextual representation.

~\cite{bahurmuz2022arabic} conducted a large-scale study combining MSA and dialectal tweets and news data. Their results show that MARBERT significantly outperforms AraBERT and other models, highlighting the importance of dialectal pre-training for social media rumour detection.

\subsection{Algerian Dialect-Specific Approaches}
Recent studies have extended rumour detection to Arabic dialects and code-switched content. ~\cite{righi2022rumor} investigated stance classification in Algerian YouTube comments using AraBERT, mBERT, and XLM-R. AraBERT achieved the best performance (macro F1 = 0.53), showing strong adaptation to dialectal data despite being trained on MSA.

Algerian dialect remains one of the least explored languages in rumour detection. ~\cite{bousri2022rumor} introduced a deep learning framework for Algerian Arabizi rumour detection using YouTube comments annotated at event and stance levels. They combined Word2Vec, ELMo, and TF-IDF representations with LSTM and GRU models enhanced by attention mechanisms, achieving an F1-score of 73\%. They also showed that stance and association features significantly improve performance, particularly in low-resource settings.

~\cite{zakaria2023algerian} focused on Algerian Arabizi using structured feature engineering. They introduced lexical, stance, and emotion-based TF-IDF representations and demonstrated that stance information is more discriminative than emotion for rumour detection. Their findings support the importance of auxiliary linguistic signals such as SDQC-style annotations.

~\cite{abdedaiem2024fassila} introduced FASSILA, a benchmark dataset for Algerian dialect fake news and sentiment analysis. They evaluated transformer models such as AraBERT, MARBERT, and DziriBERT. MARBERT achieved the best performance on fake news detection, confirming that large-scale dialectal pre-training is more effective than dialect-specific but smaller corpora.

~\cite{hamadouche2024detection} proposed a hybrid ensemble combining TF-IDF-based classical models with fine-tuned Arabic-BERT, achieving 97.72\% accuracy. However, their Algerian subset was derived from translated MSA content, limiting its realism for native dialect applications.

\subsection{Discussion and Summary}
Across the literature, several consistent trends emerge. Transformer-based models generally outperform classical and deep learning approaches, particularly when pre-trained on social media or dialectal data. MARBERT remains the most robust model for dialectal Arabic, often surpassing models trained specifically on smaller dialect corpora.

However, important limitations persist. Most existing studies focus on MSA or mixed Arabic, while Algerian dialect—especially in Arabizi form remains less explored. Dataset sizes are generally small, limiting the effectiveness of deep learning approaches. Moreover, many works rely on curated or translated datasets rather than naturally occurring social media content, reducing their validity.

These limitations highlight the need for dedicated datasets and robust framework tailored specifically to Algerian dialect social media content, which is the focus of this work.

\begin{table}[ht]
\centering
\renewcommand{\arraystretch}{1.3}
\begin{tabularx}{\textwidth}{|X|X|X|c|X|X|X|}
\hline
\textbf{Ref.} & \textbf{Dialect} & \textbf{Source} & \textbf{Size} & \textbf{Classes} & \textbf{Classifier} & \textbf{Results} \\
\hline

\cite{alkhair2019arabic} & MSA + DA & Web & 4,079 & Real / Fake & SVM & Acc: 94\%-96\% \\
\hline

\cite{zakaria2023algerian} & Algerian & YouTube & 11,822 & Rumour / Non-rumour & SVM + Stance & Not reported \\
\hline

\cite{hocini2024detecting} & Mixed & BOUTEF & 3,666 & Fake / FakeComment / NoFake & MLP & Acc: 82.3\% \\
\hline

\cite{bousri2022rumor} & Algerian & YouTube & 9,500 & Rumour / Non-rumour & Word2Vec + LSTM & F1: 73\% \\
\hline

\cite{abdedaiem2024fassila} & Algerian & YT / FB / GPT-4 & 10,087 & Real / Fake & MARBERT & Best on FND \\
\hline

\cite{hamadouche2024detection} & Arabic + DA & News & 3,563 & Real / Fake & AraBERT + Ensemble & Acc: 97.72\%, F1: 98\% \\
\hline

\cite{righi2022rumor} & Algerian & YouTube & 3,147 & Support / Deny / Query / Comment & AraBERT & F1: 0.53 \\
\hline

\cite{bahurmuz2022arabic} & MSA + DA & Twitter / News & 36,308 & Rumour / Non-rumour & MARBERT & Acc: 0.97, F1: 97\% \\
\hline

\end{tabularx}
\caption{Summary of related works on rumour and fake news detection in Arabic and Algerian dialect}
\label{tab:related_work}
\end{table}

\section{Our Proposed Approach}
\begin{figure}[h!] 
    \centering
    \includegraphics[width=\textwidth]{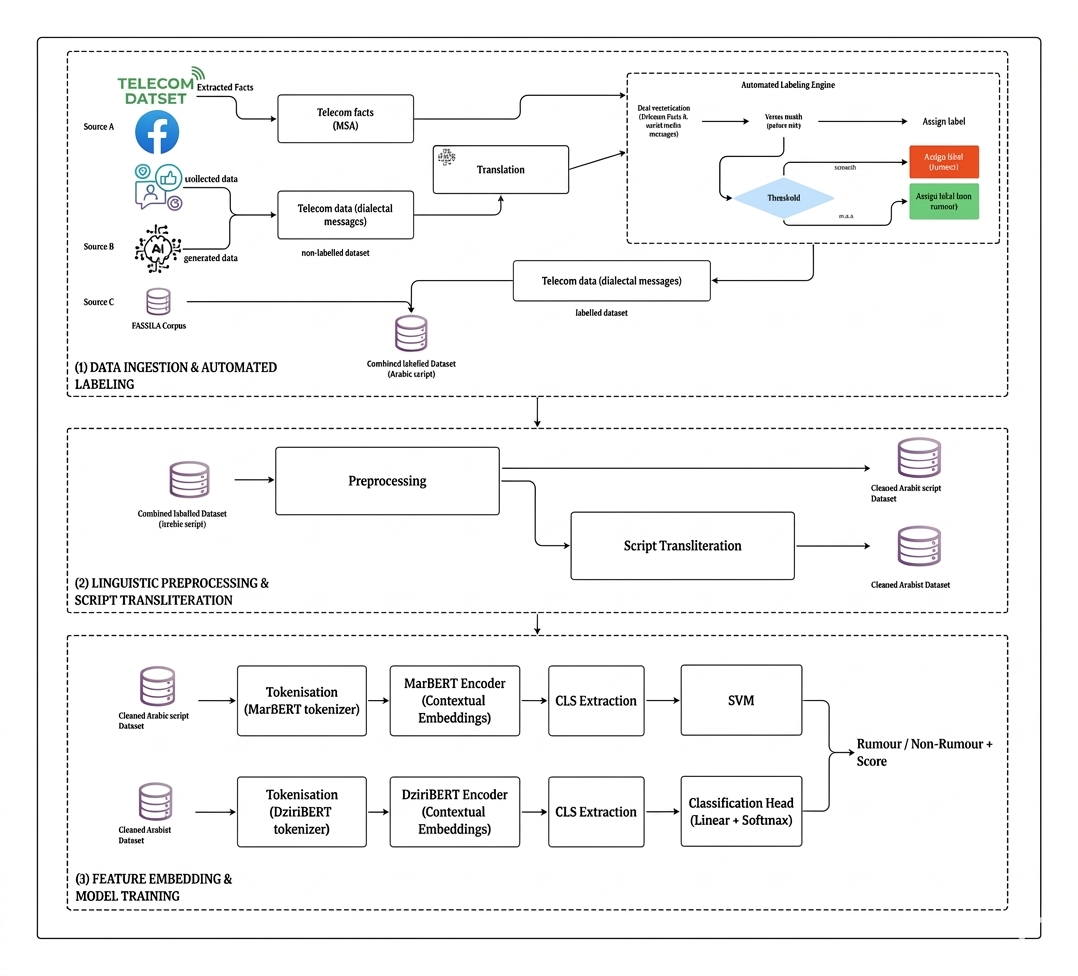} 
    \caption{Complete Framework Archietcure}
    \label{fig:architecture_systeme}
\end{figure}

The proposed framework is designed as a sequential multi-stage pipeline composed of three main modules: (i) Data Ingestion and Automated Labeling, (ii) Linguistic Preprocessing and Script Transliteration, and (iii) Feature Embedding and Model Training. The overall architecture is illustrated in Figure \ref{fig:architecture_systeme}.

\subsection{System Overview}
The objective of the proposed framework is to perform automated rumour detection in Algerian dialect telecom-related text. The system integrates heterogeneous data sources, constructs a weakly-supervised labeling mechanism based on semantic similarity, and leverages script-specific representation learning through transformer-based models.

\subsection{Module 1: Data Ingestion and Automated Labeling}
\subsubsection{Data Sources}
Due to the absence of publicly available telecom-domain rumour datasets in Algerian dialect, a custom dataset was constructed from three complementary sources:

\begin{itemize}
\item \textbf{Source A: Collected Telecom Messages} — Unlabelled user-generated content extracted from Algerian social media platforms (primarily Facebook), covering topics such as network outages, billing, and service quality. Others statements were also collected from verified Facebook pages and the official website to serve as factual references.

\item \textbf{Source B: Synthetic Messages} — Artificially generated telecom messages produced using Google Gemini to mitigate data scarcity. Prompts enforced dialectal Algerian language usage and controlled rumour/non-rumour semantics.

\item \textbf{Source C: FASSILA Corpus ~\cite{abdedaiem2024fassila}} — A pre-labelled Algerian dialect dataset [31], originally annotated for veracity. Labels were remapped to a binary scheme (rumour / non-rumour) and directly integrated into the final dataset.
\end{itemize}

\subsubsection{Automated Labeling Engine}
Since Sources A and B are unlabelled, an automated labeling mechanism was introduced based on semantic similarity with official telecom collected statements.

Each message is first translated from Algerian dialect to Modern Standard Arabic (MSA) using a large language model to enable reliable semantic alignment. Let $M = \{m_j\}_{j=1}^{k}$ denote translated messages and $F = \{f_i\}_{i=1}^{n}$ denote official facts, each associated with a validity interval $[t_i^{start}, t_i^{end}]$.

To ensure temporal consistency, only facts valid at the time of message publication $d_j$ are considered:
\begin{equation}
F(j) = \{ f_i \in F \mid t_i^{start} \le d_j \le t_i^{end} \}.
\end{equation}

If $F(j)$ is empty, all facts are used as fallback references.

Both messages and facts are encoded using the \texttt{paraphrase-multilingual-mpnet-base-v2} sentence embedding model. Cosine similarity is computed as:
\begin{equation}
\mathrm{sim}(m_j, f_i) = \frac{v_{m_j} \cdot v_{f_i}}{\|v_{m_j}\|\|v_{f_i}\|}.
\end{equation}

The final similarity score is defined as:
\begin{equation}
s_j = \max_{f_i \in F(j)} \mathrm{sim}(m_j, f_i).
\end{equation}

\subsubsection{Three-Zone Labeling Strategy}
A dual-threshold mechanism is employed to improve labeling robustness:
\begin{equation}
\text{label}(m_j) =
\begin{cases}
\text{non-rumour}, & s_j > \theta_{high} \\
\text{rumour}, & s_j < \theta_{low} \\
\text{manual review}, & \theta_{low} \le s_j \le \theta_{high}.
\end{cases}
\end{equation}

This strategy avoids overconfident classification in ambiguous cases. Samples in the intermediate region are manually validated. The final dataset is obtained by merging automatically and manually labeled samples, followed by stratified splitting into training, validation, and test sets.

\subsection{Module 2: Linguistic Preprocessing and Script Transliteration}
\subsubsection{Preprocessing Pipeline}
Given the noisy nature of Algerian dialect social media text, a normalization pipeline is applied prior to model training. The pipeline includes:

diacritics removal, Arabic letter normalization, removal of Latin fragments, punctuation and emoji filtering, elimination of metadata elements (e.g., user mentions, retweet markers), digit removal, character elongation reduction, and whitespace normalization.

This produces a cleaned Arabic-script dataset.

\subsubsection{Transliteration to Arabizi}

To support models trained on Latin-script Algerian dialect, a rule-based transliteration function (\texttt{algerian\_master\_mapper}) is applied to convert Arabic script into Arabizi.

Unlike generic transliteration systems, the proposed mapper integrates dialect-specific conventions and a protected vocabulary dictionary to preserve domain terms (e.g., \textit{mobilis}, \textit{el chabaka}, \textit{el tatbi9}).

Character-level mapping follows standard Algerian Arabizi conventions (e.g., \textit{3} for \textit{ain}, \textit{7} for \textit{ha}, \textit{9} for \textit{kaf}). The resulting output is a parallel dataset consisting of Arabic-script and Arabizi versions of each message.

\subsection{Module 3: Feature Embedding and Model Training}
\subsubsection{Model Selection}

Two transformer-based architectures are employed, selected according to script compatibility and pre-training distribution:

\begin{itemize}
\item \textbf{MarBERT} is used for Arabic-script text due to its pre-training on 128M Arabic tweets, making it well-suited for dialectal and informal content.
\item \textbf{DziriBERT} is used for Arabizi text as it is specifically trained on Algerian dialect written in Latin script.
\end{itemize}

\subsubsection{Classification Pipelines}
For both tracks, input text is tokenized and passed through the respective transformer encoder. The final hidden state corresponding to the \texttt{[CLS]} token is used as a global sentence representation.

\paragraph{Arabizi Track}
The CLS representation is passed to a linear classification head followed by a softmax layer. The model is fine-tuned end-to-end using binary cross-entropy loss:
\begin{equation}
\mathcal{L} = -\frac{1}{N} \sum_{i=1}^{N} \left[y_i \log(\hat{y}_i) + (1-y_i)\log(1-\hat{y}_i)\right].
\end{equation}

where $N$ is the number of training examples, $y_i \in \{0, 1\}$ is the true label, and $\hat{y}_i$ is the predicted probability.

\paragraph{Arabic Script Track}
A hybrid strategy is adopted. The MarBERT encoder is frozen and used as a feature extractor, and the resulting CLS embeddings are fed into a Support Vector Machine (SVM) classifier. This decouples representation learning from classification and reduces overfitting on the relatively small dataset.

\subsubsection{Baseline Models}
To ensure a comprehensive evaluation, several baselines are considered, including TF-IDF with classical machine learning models, FastText/AraVec embeddings with BiLSTM and BiGRU architectures, hybrid transformer-feature classifiers, and Llama3 evaluated in both zero-shot and fine-tuned settings.

\section{Experiments and Results}
\subsection{Dataset Construction and Statistics}
Since no publicly available dataset exists for telecom-related rumour detection in Algerian dialect, a new corpus was constructed following the methodology described in Section 3. The final dataset combines three complementary sources: (i) dialectal telecom messages collected from Algerian social media platforms, (ii) synthetic telecom messages generated using Google Gemini, and (iii) the FASSILA corpus \cite{abdedaiem2024fassila}. Table~\ref{tab:data_sources} summarizes the contribution of each source.

\begin{table}[ht]
\centering
\caption{Sources contributing to the final dataset.}
\label{tab:data_sources}
\begin{tabular}{|l|l|c|c|}
\hline
\textbf{Source} & \textbf{Description} & \textbf{Size} & \textbf{Labeling} \\
A & Collected telecom messages & 3,000 & Automated + Manual \\
B & Gemini-generated messages & 4,000 & Automated + Manual \\
C & FASSILA corpus & 4,962 & Pre-labelled \\
\textbf{Total} & & \textbf{11,962} &  \\
\end{tabular}
\end{table}

\subsubsection{Threshold Selection for Automated Labeling}
The automated labeling engine assigns veracity labels to unannotated messages based on their semantic similarity to verified telecom facts. To determine suitable decision boundaries, multiple threshold configurations were evaluated on a manually annotated validation subset.

Higher similarity thresholds ($0.7$--$0.8$) resulted in excessive rejection rates because dialectal user-generated content often exhibits substantial lexical divergence from official telecom statements written in Modern Standard Arabic. Based on empirical evaluation, the thresholds were fixed to:

[
$\theta$\_{low}=0.3, \qquad $\theta$\_{high}=0.5.
]

The resulting decision rule is defined as:

\begin{equation}
\text{label}(m_j)=
\begin{cases}
\text{non-rumour}, & s_j > \theta_{high} \\
\text{manual review}, & \theta_{low} \le s_j \le \theta_{high} \\
\text{rumour}, & s_j < \theta_{low}
\end{cases}
\label{eq:labeling_rule}
\end{equation}

Messages falling within the uncertainty interval $[0.3,0.5]$ were manually reviewed by comparing the original dialectal message, its MSA translation, and the closest matched official fact. Annotation decisions were based on factual verifiability, temporal consistency, and the distinction between factual claims and subjective opinions.

\subsubsection{Arabic-Script Dataset}
After preprocessing and normalization, the Arabic-script corpus contains 11,962 labeled messages. The dataset exhibits a relatively balanced distribution between rumour and non-rumour classes, which helps reduce classification bias during training.

Table~\ref{tab:arabic_split} reports the stratified train-validation-test partition used throughout the experiments.

\begin{table}[ht]
\centering
\caption{Arabic-script dataset split.}
\label{tab:arabic_split}
\begin{tabular}{|l|c|c|c|}
\hline
\textbf{Split} & \textbf{Total} & \textbf{Rumour} & \textbf{Non-Rumour} \\
\hline
Training & 10,149 & 4,718 & 5,431 \\
Validation & 871 & 337 & 534 \\
Test & 942 & 487 & 455 \\
\hline
Total & 11,962 & 5,542 & 6,420 \\
\hline
\end{tabular}
\end{table}

\subsubsection{Arabizi Dataset}
To support Latin-script Algerian dialect models, the Arabic-script corpus was transliterated using the proposed \textit{algerian\_master\_mapperpipeline}. The transliteration process preserves semantic content while converting Arabic orthography into conventional Algerian Arabizi.

A lightweight augmentation procedure was subsequently applied to improve robustness against spelling variation commonly observed in informal online communication. The resulting Arabizi corpus contains 12,749 messages.

Table~\ref{tab:arabizi_split} presents the final dataset split used for training and evaluation.

\begin{table}[ht]
\centering
\caption{Arabizi dataset split.}
\label{tab:arabizi_split}
\begin{tabular}{|l|c|c|c|}
\hline
\textbf{Split} & \textbf{Total} & \textbf{Rumour} & \textbf{Non-Rumour} \\
\hline
Training & 10,936 & 5,342 & 5,594 \\
Validation & 871 & 381 & 490 \\
Test & 942 & 388 & 554 \\
\hline
Total & 12,749 & 6,111 & 6,638 \\
\hline
\end{tabular}
\end{table}

\subsection{Results on the Arabizi Dataset}
Table~\ref{tab:arabizi_results} reports the parameters of all evaluated models on the Arabizi dataset.

\begin{table}[htbp]
\centering
\caption{Training configurations, Arabizi dataset.}
\label{tab:arabizi_results}
\vspace{0.2cm}
\begin{tabular}{|l|c|c|c|l|}
\hline
\textbf{Model} & \textbf{Epochs} & \textbf{Batch size} & \textbf{Learning rate} & \textbf{Other} \\ \hline
SVM + TF-IDF & -- & -- & -- & C = 1 \\ 
LR + TF-IDF & -- & -- & -- & C = 10, solver = lbfgs \\ 
RF + TF-IDF & -- & -- & -- & n\_estimators = 300 \\ \hline
BiLSTM + FastText & 20 & 32 & 1e-3 & dim = 128, patience = 4 \\ 
BiGRU + FastText & 20 & 32 & 1e-3 & dim = 128, patience = 4 \\ \hline
DziriBERT & 10 & 32 & enc: 1e-5 / clf: 5e-5 & max len = 128 \\ 
mBERT & 8 & 32 & enc: 1e-5 / clf: 5e-5 & max len = 128 \\ \hline
DziriBERT + SVM & -- & -- & -- & C = 0.01, kernel = linear \\ 
DziriBERT + LR & -- & -- & -- & C = 1, solver = lbfgs \\ 
DziriBERT + RF & -- & -- & -- & n\_estimators = 200 \\ \hline
Llama3 (FT) & 5 & 8 & 1e-4 & grad. accum. = 4, bf16 \\ 
Llama3 (zero-shot) & -- & -- & -- & temp = 0, Groq API \\ \hline
\end{tabular}
\end{table}

While the table ~\ref{tab:results01} compares the results of different tested models.

\begin{table}[htbp]
\centering
\caption{Results on the Arabizi dataset.}
\label{tab:results01}
\begin{tabular}{|l|l|c|c|c|c|}
\hline
\textbf{Family} & \textbf{Model} & \textbf{Acc.} & \textbf{Prec.} & \textbf{Rec.} & \textbf{F1} \\ \hline
\multirow{3}{*}{Classical ML} & SVM + TF-IDF & 0.77 & 0.77 & 0.77 & 0.77 \\ 
 & LR + TF-IDF & 0.76 & 0.76 & 0.76 & 0.76 \\ 
 & RF + TF-IDF & 0.76 & 0.76 & 0.76 & 0.76 \\ \hline
\multirow{2}{*}{Deep Learning} & BiLSTM + FastText & 0.76 & 0.76 & 0.75 & 0.75 \\ 
 & BiGRU + FastText & 0.79 & 0.79 & 0.79 & 0.79 \\ \hline
\multirow{2}{*}{Transformers} & mBERT & 0.80 & 0.81 & 0.80 & 0.80 \\ 
 & DziriBERT FT & \textbf{0.84} & \textbf{0.85} & \textbf{0.84} & \textbf{0.84} \\ \hline
\multirow{3}{*}{Hybrid} & DziriBERT + SVM & 0.83 & 0.83 & 0.83 & 0.83 \\ 
 & DziriBERT + LR & 0.84 & 0.84 & 0.84 & 0.84 \\ 
 & DziriBERT + RF & \textbf{0.84} & \textbf{0.85} & \textbf{0.84} & \textbf{0.84} \\ \hline
\multirow{2}{*}{LLM} & Llama3 (fine-tuned) & 0.72 & 0.72 & 0.72 & 0.72 \\ 
 & Llama3 (zero-shot) & 0.51 & 0.56 & 0.52 & 0.43 \\ \hline
\end{tabular}
\end{table}

A clear performance hierarchy emerges across model families. Classical machine learning approaches based on TF-IDF representations achieve competitive baseline performance, with SVM obtaining the highest score among traditional classifiers (F1 = 0.77). Sequence models provide moderate improvements, with BiGRU reaching F1 = 0.79, indicating that modelling word order contributes additional discriminative information beyond lexical frequencies.

Transformer-based models achieve the strongest overall performance. The multilingual baseline mBERT reaches F1 = 0.80, whereas DziriBERT, which was pre-trained specifically on Algerian dialect social media content written in Latin script, improves the score to F1 = 0.84. This result highlights the importance of matching the pre-training distribution to the target domain.

Hybrid approaches combining DziriBERT embeddings with conventional classifiers achieve performance comparable to full fine-tuning. Both DziriBERT+LR and DziriBERT+RF obtain F1 = 0.84, demonstrating that contextual embeddings extracted by DziriBERT are highly discriminative even when used as fixed features.

In contrast, Llama3 exhibits substantially weaker performance. The fine-tuned model reaches F1 = 0.72, while the zero-shot configuration drops to F1 = 0.43. These results suggest that general-purpose large language models struggle to capture the linguistic characteristics of Algerian Arabizi without extensive task-specific adaptation.

The validation loss reaches its minimum around the third epoch, whereas the validation F1 score stabilises after the fifth epoch. Although mild overfitting appears in later epochs, the final test performance indicates good generalization on unseen samples.

\subsection{Results on the Arabic-Script Dataset}
Table~\ref{tab:arabic_results} presents the parameters of each tested model on arabic Dataset.

\begin{table}[htbp]
\centering
\caption{Training configurations, Arabic script dataset.}
\label{tab:arabic_results}
\begin{tabular}{|l|c|c|c|l|}
\hline
\textbf{Model} & \textbf{Epochs} & \textbf{Batch size} & \textbf{Learning rate} & \textbf{Other} \\ \hline
SVM + TF-IDF & -- & -- & -- & C = 1 \\ 
LR + TF-IDF & -- & -- & -- & C = 5, solver = liblinear \\ 
RF + TF-IDF & -- & -- & -- & n\_estimators = 200 \\ \hline
BiLSTM + FastText & 20 & 32 & 1e-3 & dim = 200, patience = 4 \\ 
BiGRU + FastText & 20 & 32 & 1e-3 & dim = 200, patience = 4 \\ 
AraVec + BiLSTM & 20 & 32 & 1e-3 & dim = 300, patience = 4 \\ \hline
AraBERT & 8 & 16 & enc: 2e-5 / clf: 1e-4 & max len = 128 \\ 
DziriBERT & 5 & 8 & enc: 2e-5 / clf: 1e-4 & max len = 128 \\ 
GigaBERT & 8 & 16 & enc: 2e-5 / clf: 1e-4 & max len = 128 \\ 
MarBERT & 8 & 16 & enc: 2e-5 / clf: 1e-4 & max len = 128 \\ 
mBERT & 8 & 32 & enc: 1e-5 / clf: 5e-5 & max len = 128 \\ \hline
MarBERT + SVM & -- & -- & -- & C = 0.01, kernel = linear \\ 
MarBERT + LR & -- & -- & -- & C = 1, solver = lbfgs \\ 
MarBERT + RF & -- & -- & -- & n\_estimators = 200 \\ \hline
\end{tabular}
\end{table}

Table~\ref{tab:arabic_scores} summarizes the results obtained on the Arabic-script dataset.

\begin{table}[htbp]
\centering
\caption{Results on the Arabic script dataset.}
\label{tab:arabic_scores}
\begin{tabular}{|l|l|c|c|c|c|}
\hline
\textbf{Family} & \textbf{Model} & \textbf{Acc.} & \textbf{Prec.} & \textbf{Rec.} & \textbf{F1} \\ \hline
\multirow{3}{*}{Classical ML} 
 & SVM + TF-IDF & 0.77 & 0.76 & 0.77 & 0.76 \\
 & LR + TF-IDF  & 0.77 & 0.76 & 0.77 & 0.76 \\
 & RF + TF-IDF  & 0.75 & 0.75 & 0.75 & 0.75 \\ \hline
\multirow{3}{*}{Deep Learning} 
 & BiLSTM + FastText & 0.73 & 0.73 & 0.73 & 0.73 \\
 & BiGRU + FastText  & 0.73 & 0.73 & 0.73 & 0.73 \\
 & AraVec + BiLSTM   & 0.72 & 0.73 & 0.72 & 0.72 \\ \hline
\multirow{5}{*}{Transformers} 
 & mBERT        & 0.79 & 0.79 & 0.79 & 0.79 \\
 & GigaBERT     & 0.81 & 0.81 & 0.80 & 0.80 \\
 & AraBERT      & 0.81 & 0.81 & 0.82 & 0.81 \\
 & DziriBERT FT & 0.82 & 0.82 & 0.82 & 0.82 \\
 & MarBERT      & \textbf{0.83} & \textbf{0.83} & \textbf{0.82} & \textbf{0.82} \\ \hline
\multirow{3}{*}{Hybrid} 
 & MarBERT + SVM & \textbf{0.84} & \textbf{0.84} & \textbf{0.84} & \textbf{0.84} \\
 & MarBERT + LR  & 0.84 & 0.84 & 0.84 & 0.84 \\
 & MarBERT + RF  & 0.84 & 0.84 & 0.84 & 0.84 \\ \hline
\end{tabular}
\end{table}

Classical machine learning models again provide strong baselines, with SVM and Logistic Regression reaching F1 = 0.76. Interestingly, recurrent neural architectures fail to surpass these baselines. BiLSTM, BiGRU, and AraVec-based models achieve F1 scores between 0.72 and 0.73, suggesting that static word embeddings are insufficient to fully capture the linguistic variability of Algerian dialect written in Arabic script.

Transformer models consistently outperform both classical and recurrent approaches. Among the evaluated encoders, MarBERT and DziriBERT obtain the strongest standalone results (F1 = 0.82), followed by AraBERT (F1 = 0.81), GigaBERT (F1 = 0.80), and mBERT (F1 = 0.79). The superiority of MarBERT is expected given its large-scale pre-training on Arabic social media content, which closely resembles the distribution of the collected corpus.

The best overall performance is achieved by the hybrid architectures built upon MarBERT embeddings. MarBERT+SVM, MarBERT+LR, and MarBERT+RF all obtain F1 = 0.84, surpassing end-to-end transformer fine-tuning. This finding indicates that contextual representations extracted by MarBERT remain highly informative when combined with lightweight discriminative classifiers.

\section{Analysis and Discussion}
\subsection{Effect of Model Architecture}
The experimental results reveal a clear performance hierarchy across model families. Classical machine learning models based on TF-IDF representations provide strong and remarkably stable baselines, achieving F1 scores between 0.75 and 0.77 on both datasets. Their competitive performance indicates that rumour detection in the telecom domain relies partly on discriminative lexical cues that can be captured through frequency-based representations.

Recurrent neural architectures provide mixed results. On the Arabizi dataset, BiGRU improves over the TF--IDF baseline and reaches an F1 score of 0.79, suggesting that sequential modelling contributes useful contextual information. However, recurrent models consistently underperform on the Arabic-script dataset, where scores remain between 0.72 and 0.73. This behaviour highlights the limitations of static embeddings when dealing with the rich morphology of Arabic dialects.

Transformer-based models produce the largest performance gains. Across both datasets, models pre-trained on social media and dialectal content consistently outperform those trained primarily on formal Arabic corpora. DziriBERT achieves the best standalone performance on the Arabizi dataset (F1 = 0.84), while MarBERT obtains the strongest standalone result on the Arabic-script dataset (F1 = 0.82). These findings confirm that pre-training corpus alignment is a more important factor than model size or architecture alone.

The proposed hybrid approach achieves the best overall performance. Combining transformer-derived CLS embeddings with lightweight classifiers yields an F1 score of 0.84 on both datasets. The marginal difference between SVM, Logistic Regression, and Random Forest classifiers suggests that once high-quality contextual representations are available, the representation itself contains most of the discriminative information required for classification.

Large language models show comparatively weak performance. Fine-tuned Llama3 reaches F1 = 0.72, while zero-shot prompting achieves only F1 = 0.43. Despite their large parameter count, these models appear less effective than task-specific transformers, likely due to limited exposure to Algerian dialect content during pre-training and the mismatch between generative pre-training objectives and discriminative classification tasks.

\subsection{Impact of Script Representation}
A comparison between the Arabic-script and Arabizi datasets reveals that the proposed transliteration pipeline preserves most of the information required for rumour detection. Classical machine learning and transformer-based models exhibit only minor performance differences across the two tracks, typically within one or two F1 points.

The largest discrepancy appears for recurrent neural architectures, which perform substantially better on Arabizi than on Arabic script. This observation suggests that the morphological complexity of Arabic script poses a challenge for models relying on static embeddings and sequential processing. In contrast, transformer architectures largely eliminate this gap through subword tokenisation and contextual encoding mechanisms.

Overall, both script representations lead to comparable performance, with the best models achieving an identical F1 score of 0.84. This result validates the effectiveness of the proposed transliteration strategy and demonstrates that Arabizi can serve as a viable alternative representation for Algerian dialect rumour detection.

\subsection{Comparison with Existing Approaches}
To assess the effectiveness of the proposed framework, representative approaches from the literature were re-implemented and evaluated under the same experimental conditions. The results show that traditional machine learning and recurrent neural architectures remain competitive but consistently lag behind transformer-based methods.

The strongest baseline, represented by fine-tuned MarBERT, achieves an F1 score of 0.82. In contrast, the proposed approaches reach F1 = 0.84 on both script tracks. Although the improvement may appear modest in absolute terms, it is obtained on a challenging dialectal dataset and remains consistent across all evaluation metrics.

It is important to note that the original studies were developed for different domains and datasets. Consequently, the comparison should be interpreted as an evaluation of transferability to the Algerian telecom rumour detection setting rather than a direct replication of previously reported benchmarks.

\subsection{Error Analysis and Limitations}
Analysis of the confusion matrices reveals that both DziriBERT and MarBERT produce substantially more false negatives than false positives. In practical terms, the models are more likely to miss a rumour than to incorrectly flag legitimate information. This behaviour suggests that many rumours are expressed through subtle paraphrases or indirect formulations that differ from patterns observed during training.

The Arabic-script track exhibits a higher false-negative rate than the Arabizi track, which is consistent with the greater morphological variability of Arabic text. Prefixation, suffixation, and derivational variation often produce semantically equivalent rumours with significantly different surface forms.

Several limitations should also be acknowledged. First, part of the dataset was generated synthetically using a large language model, which may not perfectly reflect the distribution of naturally occurring rumours. Second, the automated labeling process relies on similarity matching against official telecom facts and may occasionally assign incorrect labels to ambiguous cases. Third, the dataset focuses exclusively on the Algerian telecom domain, limiting the direct generalisability of the results to other sectors. Finally, all experiments were conducted on a single dialect, and future work should investigate the transferability of the proposed framework to other Arabic dialects and multilingual misinformation settings.

\section{Conclusion}
This paper addressed automatic rumour detection in Algerian dialect social media content within the telecommunications domain. The main challenge lies in the dual-script nature of Algerian online communication (Arabic script and Arabizi) and the scarcity of annotated datasets for this setting.

To overcome these limitations, we proposed an end-to-end pipeline covering dataset construction, preprocessing and transliteration, and classification. A dedicated dataset was built by combining real social media data, synthetic samples, and the FASSILA corpus, with automated similarity-based annotation. Extensive experiments were conducted across multiple model families and both script representations.

Results show that domain-specific pretraining is more effective than model scale in low-resource dialect settings. The best configurations achieve an F1-score of 0.84 on both Arabizi and Arabic tracks, outperforming existing approaches by a significant margin.

Despite these results, limitations remain regarding dataset generalization, annotation noise, and limited use of contextual signals. Future work will focus on expanding the dataset across domains, integrating richer contextual features, and exploring more advanced multilingual and dialect-aware models, with the goal of improving robustness and real-world applicability.

\bibliographystyle{unsrtnat}
\bibliography{references} 
\end{document}